\documentclass{article}
\usepackage{spconf,amsmath,amssymb,amsthm,graphicx,caption}
\usepackage[ruled,vlined,linesnumbered,resetcount]{algorithm2e}
\usepackage{enumitem}
\usepackage{cancel}
\usepackage{subcaption}
\usepackage{balance}

\theoremstyle{definition}

\newtheorem{remark}{Remark}

\title{Robust Principal Component Analysis Based On Maximum Correntropy\\ Power Iterations}
\name{Jean P. Chereau, Bruno Scalzo Dees, Danilo P. Mandic}
\address{Department of EEE, Imperial College London, London, SW7 2BT, UK \\
	Emails: \{jean.chereau18, bruno.scalzo-dees12, d.mandic\}@imperial.ac.uk
}
\begin{document}
\ninept
\maketitle
\begin{abstract}
Principal component analysis (PCA) is recognised as a quintessential data analysis technique when it comes to describing linear relationships between the features of a dataset. However, the well-known sensitivity of PCA to non-Gaussian samples and/or outliers often makes it unreliable in practice. To this end, a robust formulation of PCA is derived based on the maximum correntropy criterion (MCC) so as to maximise the expected likelihood of Gaussian distributed reconstruction errors. In this way, the proposed solution reduces to a generalised power iteration, whereby: (i) robust estimates of the principal components are obtained even in the presence of outliers; (ii) the number of principal components need not be specified in advance; and (iii) the \textit{entire} set of principal components can be obtained, unlike existing approaches. The advantages of the proposed maximum correntropy power iteration (MCPI) are demonstrated through an intuitive numerical example.
\end{abstract}
\section{Introduction}
\label{sec:intro}

The principal component analysis (PCA) method \cite{pca1}, \cite{pca2} is not only well-established but has become a standard data analysis technique for explaining the relationships between latent features in a dataset. Despite its physical interpretability and practicality, owing to its linear nature of modelling component dependence, PCA is only truly optimal when its reconstruction error is distributed according to a multivariate Gaussian distribution. This can be seen from the minimum square error formulation of PCA, given by
\begin{align} \label{eq:LS_PCA}
	\min_{\bf \Lambda, V} \quad & \| {\bf X}^{T}{\bf X} - {\bf V}{\bf \Lambda}{\bf V}^{T} \|_{2}^{2} \\
	\textnormal{s.t.} \quad & {\bf V}^{T}{\bf V}={\bf I} \notag
\end{align}
where ${\bf X} = [{\bf x}_{1}, ..., {\bf x}_{n}]^{T} \in \mathbb{R}^{n \times p}$ denotes the data matrix which contains as its rows $n$ samples of the variable, ${\bf x}_{k} \in \mathbb{R}^{p}$, for $k=1,...,n$, and ${\bf \Lambda}\in  \mathbb{R}^{p \times p}$ and ${\bf V} \in  \mathbb{R}^{p \times p}$ are respectively the matrices containing the principal variances and their associated components. By virtue of the above formulation, the results obtained from PCA are highly sensitive to outliers and/or any general form of non-Gaussian behaviour; this is reflected in the distribution of the reconstruction error samples.

To overcome this issue, several extensions of PCA have been proposed to cater for non-linear and/or non-Gaussian relationships. This includes the use of kernel functions, which serve as a building block to describe any form of non-linear and/or non-Gaussian relationship between features \cite{rpca1}, \cite{rpca2}, \cite{rpca3}. This class of extensions is referred to as \textit{kernel PCA} \cite{kpca}.  Despite its robust performance in practice, the use of PCA remains more widespread than that of kernel PCA, notably because of its easy implementation and interpretability in terms of linear relationships. 

More recently, the so-called \textit{correntropy} measure \cite{Principe2006,correntropy2,Principe2007} has been successfully employed as a kernel function to deal with non-linear principal components \cite{kpcacorr}. Correntropy can be regraded as a natural higher order extension of the standard, second order, correlation measure. In this way, it addresses the problem that most of the conventional information theoretic learning measures do not account for; that of higher order moments of the probability distribution function of the data \cite{Principe2000}. This measure has served as an efficient tool for analysing higher order statistical moments in non-Gaussian signals \cite{Principe2007, Principe2009, Principe2015}, and has even found its applications as a cost function for robust optimization, adaptive filtering, and machine learning, within the \textit{maximum correntropy criterion} (MCC) paradigm \cite{Principe2009_2,Principe2007}.

An MCC-based formulation of PCA has been proposed in \cite{mccpca1} and has shown significant potential for robust dimensionality reduction. However, there remain several issues that need to be addressed prior to its more widespread application. Namely, the framework in \cite{mccpca1}, which is based on the half-quadratic optimization algorithm, separates the data into two subspaces: (i) the \textit{feature space}, which is spanned by the $m \geq 1$ first principal components; and (ii) the \textit{noise space}, which is spanned by the remaining principal components. This requires the user to pre-specify the number of principal components to include in the feature space. In addition, although the feature space and the noise space are separated in a robust manner, the principal components within each distinct subspace are calculated through standard PCA. In this way, if the procedure in \cite{mccpca1} were applied under the assumption that the feature space be spanned by the \textit{entire} set of principal components, then the solution would reduce to standard PCA.

To this end, we introduce a novel, more flexible, formulation for PCA based on the the maximum correntropy criterion. This is achieved by employing a generalised form of the power iterations method \cite{Journee2010} to maximise the MCC reward function. This allows for not only the number of principal components not to be specified in advance, but also for the \textit{entire} set of principal components to be estimated, unlike the existing version in \cite{mccpca1}. The advantages of the proposed maximum correntropy power iteration for robust and flexible PCA, even in the presence of outliers, are demonstrated through an intuitive numerical example.

\section{Maximum Correntropy Criterion}
\label{sec:mcc}

The measure of \textit{correntropy} between two $p$-dimensional random variables, ${\bf x}, {\bf y} \in \mathbb{R}^{p}$, can be interpreted probabilistically through the likelihood of ${\bf x} = {\bf y}$. Equivalently, by defining the difference (or error) between the random variables as ${\bf e} \equiv ({\bf x} - {\bf y})$, the measure of correntropy reduces to the likelihood of the event ${\bf e} = {\bf 0}$. Under the assumption that ${\bf e} \in \mathbb{R}^{p}$ is multivariate Gaussian distributed, the correntropy between ${\bf x}$ and ${\bf y}$ is defined as
\begin{equation} \label{eq:corr1}
\kappa_{\mathbf{\Sigma}}(\mathbf{x}, \mathbf{y}) = \mathrm{exp}\left(-\frac{1}{2}(\mathbf{x} - \mathbf{y})^{T}\mathbf{\Sigma}^{-1}(\mathbf{x} - \mathbf{y})\right)
\end{equation}
where $\mathbf{\Sigma}$ is a positive definite matrix, which can be thought of as the \textit{a priori} covariance matrix of the variable ${\bf e} \in \mathbb{R}^{p}$ in the ideal Gaussian setting (in the absence of outliers and/or non-Gaussian samples). For the particular case where $\mathbf{\Sigma}$ is assumed to be spherical (or isotropic), that is, $\mathbf{\Sigma} \equiv \sigma^{2}\mathbf{I}$, the relation in (\ref{eq:corr1}) can be rewritten as
\begin{equation}
\kappa_{\sigma}(\mathbf{x}, \mathbf{y}) =  \mathrm{exp}\left({-\frac{\lVert\mathbf{x} - \mathbf{y}\rVert_{2}^{2}}{2\sigma^{2}}}\right)
\end{equation}
where the parameter $\sigma$ is referred to as the \textit{kernel size}. For ease of notation, we use $\kappa_{\sigma}(\mathbf{x} - \mathbf{y}) \equiv \kappa_{\sigma}(\mathbf{x}, \mathbf{y})$. 

Next, consider $n$ samples of the random variables ${\bf x}_{k}, {\bf y}_{k} \in \mathbb{R}^{p}$, for $k=1,...,n$, where ${\bf y}_{k} = \mathbf{f}(\mathbf{x}_{k}; \mathbf{v})$ is given by the output of a general non-linear model, with ${\bf v} \in \mathbb{R}^{p}$ being the variable of interest. The maximum correntropy criterion then asserts that, for a fixed kernel size $\sigma$, the variable ${\bf v}$ can be chosen so as to maximise the \textit{expected correntropy} between ${\bf y}_{k}$ and ${\bf x}_{k}$, thereby maximising the expected likelihood that the error vector vanishes, i.e. ${\bf e}_{k} = ({\bf y}_{k} - \mathbf{f}(\mathbf{x}_{k}; \mathbf{v})) = {\bf 0}$. This can be formulated as the following maximization problem
\begin{equation}
    \max_{\bf v} \;\; V_{\sigma} = E \{ \kappa_{\mathbf{\sigma}}(\mathbf{x}_{k}, \mathbf{y}_{k}) \} =  \frac{1}{n}\sum_{k=1}^{n}\kappa_{\sigma}\left({\bf y}_{k} - \mathbf{f}(\mathbf{x}_{k}; \mathbf{v})\right)
\end{equation}
where $V_{\sigma}$ denotes an appropriate Parzen estimator \cite{Parzen1962}. The parameter ${\bf v}$ which attains the maximum value of the MCC reward function can therefore be obtained by inspecting the stationary point, $\frac{\partial V_{\sigma}}{\partial {\bf v}} = {\bf 0}$, and solving for ${\bf v}$ accordingly.

\begin{remark}
	While the minimisation of the $L_{2}$-norm of errors (least squares) \textit{penalizes error} and is thus sensitive to large errors (outliers), the correntropy maximisation \textit{rewards the goodness of fit}, thereby being sensitive to small errors only and hence robust to outliers.
\end{remark}



\section{Maximum Correntropy Power Iterations}
\label{sec:theory}

\subsection{First Principal Component}
\label{subsec:solution}

Following the least squares formulation of PCA in (\ref{eq:LS_PCA}), the procedure for determining the first principal component can be thought of as that of finding the vector $\mathbf{v}_{1} \in \mathbb{R}^{p}$ which minimises the distance, in the Euclidean sense, between the samples $\mathbf{x}_{k}$ and their respective projections on the space spanned by $\mathbf{v}_{1}$, that is
\begin{align}
\min_{{\bf v}_{1}} \quad & \sum_{k=1}^{n}\left\lVert(\mathbf{I} - \mathbf{v}_{1}\mathbf{v}_{1}^{T})\mathbf{x}_{k}\right\rVert_{2}^{2} \\
\textnormal{s.t.} \quad & {\bf v}_{1}^{T}{\bf v}_{1}=1 \notag
\end{align}
This allows us to consider the PCA problem formulation under the MCC paradigm as follows. The task boils down to determining the vector $\mathbf{v}_{1}$ which maximises the expected correntropy between the samples $\mathbf{x}_{k}$ and their respective projections on the space spanned by $\mathbf{v}_{1}$. In other words, this is equivalent to maximising the expected likelihood of the event that $\mathbf{x}_{k}$ is equal to its projection on the space spanned by $\mathbf{v}_{1}$. Under the assumption that the reconstruction error is Gaussian distributed, this problem can be formally written as follows \cite{optimization}
\begin{align}
    \max_{\mathbf{v}_{1}} \quad  & \sum_{k=1}^{n}\kappa_{\sigma}\left((\mathbf{I} - \mathbf{v}_{1}\mathbf{v}_{1}^{T})\mathbf{x}_{k}\right) \\
    \textnormal{s.t.} \quad  & \mathbf{v}_{1}^{T}\mathbf{v}_{1} = 1 \notag
    \end{align}
The Lagrangian function associated with the above optimization task is then given by
\begin{equation}
    \mathcal{L}(\mathbf{v}_{1}, \lambda_{1}) = \sum_{k=1}^{n}\kappa_{\sigma}\left((\mathbf{I} - \mathbf{v}_{1}\mathbf{v}_{1}^{T})\mathbf{x}_{k}\right) - \frac{\lambda_{1}}{2\sigma^{2}}(\mathbf{v}_{1}^{T}\mathbf{v}_{1} - 1)
\end{equation}
with $\lambda_{1}$ as the Lagrange multiplier. The solution pair, $(\mathbf{v}_{1}, \lambda_{1})$, can be found through differentiation to yield the following generalised eigenvalue problem
\begin{equation} \label{eq:generalisedEVD}
    \mathbf{X}^{T}\mathbf{G}_{\sigma}\mathbf{X}\mathbf{v}_{1} = \lambda_{1}\mathbf{v}_{1}
\end{equation}
where $\mathbf{G}_{\sigma} \in \mathbb{R}^{n \times n}$ is a diagonal matrix with its $(k,k)$-th entry defined as
\begin{equation} 
    [\mathbf{G}_{\sigma}]_{k k} = \kappa_{\sigma}\left((\mathbf{I} - \mathbf{v}_{1}\mathbf{v}_{1}^{T})\mathbf{x}_{k}\right)
\end{equation}
It is important to mention that the solution in (\ref{eq:generalisedEVD}) is not given in a closed-form, as the matrix $\mathbf{G}_{\sigma}$ is a function of the variable of interest, ${\bf v}_{1}$. We therefore have resort to a fixed-point solution \cite{fixedpoint} which exhibits good convergence properties for a large enough kernel size, $\sigma$. The solution of interest is then obtained by the eigenvector associated with the \textit{largest} eigenvalue of the correntropy-weighted scatter matrix, $\mathbf{X}^{T}\mathbf{G}_{\sigma}\mathbf{X}$. 

\begin{remark}
	Based on the results in \cite{Journee2010}, any concave function, denoted by $f({\bf v})$, can be maximised with respect to the vector-valued variable, ${\bf v}$, by employing a generalised form of the power iteration \cite{pim}, given by ${\bf v} \leftarrow \frac{\partial f({\bf v})}{\partial {\bf v}} / \left\| \frac{\partial f({\bf v})}{\partial {\bf v}} \right\|$.
	Given the concave nature of the MCC, whereby $f \equiv V_{\sigma}$, the maximum correntropy power iteration can be obtained as
	\begin{equation} \label{eq:generalisedPM}
	{\bf v} \leftarrow \frac{\mathbf{X}^{T}\mathbf{G}_{\sigma}\mathbf{X}{\bf v}}{\left\| \mathbf{X}^{T}\mathbf{G}_{\sigma}\mathbf{X}{\bf v} \right\|}
	\end{equation}
	The above expression is an alternative way to arrive at the generalised eigenvalue problem in (\ref{eq:generalisedEVD}). It is important to recall that the diagonal matrix, $\mathbf{G}_{\sigma}$, is dependent of ${\bf v}$, so that $\mathbf{G}_{\sigma}$ can be updated once the power iteration scheme in (\ref{eq:generalisedPM}) converges to a solution, ${\bf v}$, for a fixed $\mathbf{G}_{\sigma}$. The repeated application of this alternating procedure yields the globally optimum value of ${\bf v}$. The proposed maximum correntropy power iteration is summarised in Algorithm 1.
\end{remark}

\vspace{-0.5cm}

\begin{algorithm}[b]
	\caption{Maximum correntropy power iteration.}
	\SetAlgoLined
	\SetKwInOut{Input}{Input}
	\SetKwInOut{Output}{Output}
	\Input{$\mathbf{X}$, $\sigma$}
	\Output{$\mathbf{v}_{1}$}
	\While{$\mathrm{not\, converged}$}{
		
		$[\mathbf{G}_{\sigma}]_{k k} \leftarrow \kappa_{\sigma}\left((\mathbf{I} - \mathbf{v}_{1}\mathbf{v}_{1}^{T})\mathbf{x}_{k}\right)$

		\While{$\mathrm{not\, converged}$}{
			$\mathbf{v}_{1} \leftarrow \frac{\mathbf{X}^{T}\mathbf{G}_{\sigma}\mathbf{X}\mathbf{v}_{1}}{\| \mathbf{X}^{T}\mathbf{G}_{\sigma}\mathbf{X}\mathbf{v}_{1} \|}$
		}
	}
\end{algorithm}

\pagebreak

\subsection{Remaining Principal Components}
\label{subsec:eig}

We next consider the estimation of the remaining principal components, ${\bf v}_{i}$, for $i>1$. Consider the case where $(i - 1) \geq 1$ principal components, $\mathbf{V}_{1:(i-1)} = [{\bf v}_{1},...,{\bf v}_{(i-1)}] \in \mathbb{R}^{p \times (i-1)}$, corresponding to the $(i - 1)$ largest variances in the data, have been obtained. In order to find the $i$-th component, it is necessary to remove (deflate) from the data the previous $(i-1)$ components. By introducing the projection matrix, $\mathbf{P}_{(i-1)} = \mathbf{V}_{1:(i-1)}\mathbf{V}_{1:(i-1)}^{T}$, the deflated data matrix with the existing principal components removed is given by $\mathbf{X}_{(i)} \equiv \mathbf{X}(\mathbf{I} - \mathbf{P}_{(i-1)})$.

Recall that the $i$-th principal component, $\mathbf{v}_{i} \in \mathbb{R}^{p}$, must satisfy the following three conditions: (i) it must be orthogonal to the first $(i-1)$ components contained in $\mathbf{V}_{1:(i-1)}$; (ii) it must be a unit vector; and (iii) it must maximise the expected correntropy between the deflated data samples, $\mathbf{x}_{(i), k} \equiv (\mathbf{I} - \mathbf{P}_{(i-1)})\mathbf{x}_{k}$, and their respective projections on the subspace spanned by $\mathbf{v}_{i}$. This problem can be formally written as
\begin{align}
    \max_{\mathbf{v}_{i}} \quad & \sum_{k=1}^{n}\kappa_{\sigma}\left((\mathbf{I} - \mathbf{P}_{(i-1)} - \mathbf{v}_{i}\mathbf{v}_{i}^{T})\mathbf{x}_{k}\right) \\
    \textnormal{s.t.} \quad & \mathbf{v}_{i}^{T}\mathbf{v}_{i} = 1 \notag\\
    & \mathbf{v}_{i}^{T}\mathbf{P}_{(i-1)}\mathbf{v}_{i} = 0 \notag
    \end{align}
with the corresponding Lagrangian function now given by
\begin{align}
    \mathcal{L}(\mathbf{v}_{i}, \lambda_{i}, \lambda_{i}') & = \sum_{k=1}^{n}\kappa_{\sigma}\left((\mathbf{I} - \mathbf{P}_{(i-1)} - \mathbf{v}_{i}\mathbf{v}_{i}^{T})\mathbf{x}_{k}\right)  \notag\\
    & - \frac{\lambda_{i}}{2\sigma^{2}}(\mathbf{v}_{i}^{T}\mathbf{v}_{i} - 1) - \frac{\lambda_{i}'}{2\sigma^{2}}\mathbf{v}_{i}^{T}\mathbf{P}_{(i-1)}\mathbf{v}_{i}
\end{align}
Upon inspecting the stationary points of the Lagrangian, we obtain the following equality
\begin{equation}
    \left(\mathbf{S} - \mathbf{P}_{(i-1)}\mathbf{S} - \mathbf{S}\mathbf{P}_{(i-1)}\right)\mathbf{v}_{i} = (\lambda_{i}\mathbf{I} + \lambda_{i}'\mathbf{P}_{(i-1)})\mathbf{v}_{i}
\end{equation}
where $\mathbf{S} \equiv \mathbf{X}^{T}\mathbf{G}_{\sigma}\mathbf{X}$ denotes the correntropy-weighted scatter matrix, with the $(k,k)$-th element of the diagonal matrix, $\mathbf{G}_{\sigma}$, given by
\begin{equation}
    [\mathbf{G}_{\sigma}]_{k k} = \kappa_{\sigma}\left((\mathbf{I} - \mathbf{P}_{(i-1)} - \mathbf{v}_{i}\mathbf{v}_{i}^{T})\mathbf{x}_{k}\right)
\end{equation}
Equation (10) can be simplified to a generalised eigenvalue problem by setting $\lambda_{i} = \lambda_{i}'$. From the perspective of the Lagrange multiplier, this choice is reasonable as it imposes equal importance on both constraints. Because the matrix sum $(\mathbf{I} + \mathbf{P}_{(i)})$ must be full-rank for $0 \leq i < p$, then it is invertible, which means that we can define the matrix $\mathbf{Q}_{(i)} \equiv (\mathbf{I} + \mathbf{P}_{(i)})^{-1} \in \mathbb{R}^{p \times p}$. This provides us with the solution pair $(\mathbf{v}_{i}, \lambda_{i})$ based on the following eigenproblem
\begin{equation}
   \mathbf{Q}_{(i-1)}\left(\mathbf{S} - \mathbf{P}_{(i-1)}\mathbf{S} - \mathbf{S}\mathbf{P}_{(i-1)}\right)\mathbf{v}_{i} = \lambda_{i}\mathbf{v}_{i}
\end{equation}
Again, this solution reduces to a generalised power iteration. Notice that, for every step $1 < i < p$, the MCC is maximised by the eigenvector associated with the \textit{largest} eigenvalue of the matrix $\mathbf{K} \equiv \mathbf{Q}_{(i-1)}\left(\mathbf{S} - \mathbf{P}_{(i-1)}\mathbf{S} - \mathbf{S}\mathbf{P}_{(i-1)}\right)$. The search for this eigenvector can be performed via a maximum correntropy power iteration as in Algorithm 1.

It is important to highlight that the maximum correntropy power iteration converges to the eigenvector associated with the largest eigenvalue in \textit{magnitude} of the matrix $\mathbf{K}$. It is not difficult to see that for $i > 0$ the matrix $\mathbf{K} \in \mathbb{R}^{p \times p}$ will be indefinite (i.e. contains both positive and negative eigenvalues). We therefore cannot exclude the possibility that the most negative eigenvalue of $\mathbf{K}$ may be larger in magnitude than the most positive eigenvalue of $\mathbf{K}$. This issue can be rectified by adding a large enough positive constant, $\theta$, to all diagonal elements of $\mathbf{K}$, that is
\begin{equation}
\left(\mathbf{K} + \theta\mathbf{I}\right)\mathbf{u} = \lambda\mathbf{u} \Leftrightarrow \mathbf{K}\mathbf{u} = (\lambda - \theta)\mathbf{u} = \lambda'\mathbf{u}
\end{equation}
whereby the eigenvectors of $\mathbf{K}$ remain unaffected. The overall solution therefore reduces to a modified maximum correntropy power iteration, described in Algorithm 2.

\subsection{Last Principal Component}
\label{subsec:lasteig}

By definition, the principal components of a dataset are orthogonal to one another, and therefore the last principal component must exist in the null space of the matrix $\mathbf{V}_{1:(p-1)}^{T} \in \mathbb{R}^{(p-1) \times p}$. Since $\mathrm{Rank}(\mathbf{V}_{1:(p-1)}) = (p-1)$, the null space is spanned by a single vector in $\mathbb{R}^{p}$. With the additional unit-norm constraint imposed on the principal component, we can conclude that $\mathbf{v}_{p} \in \mathbb{R}^{p}$ must be a unit vector which spans the null space of $\mathbf{V}_{1:(p-1)}^{T}$.


\subsection{Kernel Size Shrinking Algorithm}
\label{sec:algorithm}

Based on the deflation procedure described in Section \ref{subsec:eig}, the use of Algorithm 2 alone is sufficient to estimate the entire set of principal components. However, the convergence of the proposed maximum correntropy power iteration largely depends on the chosen kernel size, $\sigma$ \cite{Silverman1986}. This can be seen from the Taylor series expansion of the MCC reward function for determining the $i$-th principal component, given by
\begin{equation} \label{eq:Taylor}
	V_{\sigma} = \sum_{n=0}^{\infty} \frac{(-1)^{n}}{2^{n}\sigma^{2n} n!} E \left\{ \|{\bf e}_{i}\|_{2}^{2n} \right\}
\end{equation}
where ${\bf e}_{i} = (\mathbf{I} - \mathbf{v}_{i}\mathbf{v}_{i}^{T})\mathbf{x}_{k} \in \mathbb{R}^{p}$ is the reconstruction error of the $i$-th principal component. If the kernel size is too small, then the MCC reward function may contain multiple local maxima, thereby reducing the chances of the fixed-point iteration scheme reaching the global maximum. Conversely, if the kernel size is too large then the MCC reward function becomes sensitive to outliers (the quadratic term in (\ref{eq:Taylor}) dominates), and the performance of the maximum correntropy power iteration approaches that of standard PCA, as $\sigma \to \infty$ \cite{Principe2007}.

\begin{algorithm}[t]
	\SetAlgoLined
	\SetKwInOut{Input}{Input}
	\SetKwInOut{Output}{Output}
	\Input{$\mathbf{X}$, $\mathbf{P}_{(i-1)}$, $\mathbf{Q}_{(i-1)}$, $\mathbf{v}_{i}$, $\sigma_{i}$}
	\Output{$\mathbf{v}_{i}$}
	\While{$\mathrm{not\, converged}$}{
		
		$[\mathbf{G}_{\sigma_{i}}]_{k k} \leftarrow \kappa_{\sigma_{i}}\left((\mathbf{I} - \mathbf{P}_{(i-1)} - \mathbf{v}_{i}\mathbf{v}_{i}^{T})\mathbf{x}_{k}\right)$
		
		$\mathbf{S} \leftarrow \mathbf{X}^{T}\mathbf{G}_{\sigma_{i}}\mathbf{X}$
		
		$\mathbf{K} \leftarrow \mathbf{Q}_{(i-1)}(\mathbf{S} - \mathbf{P}_{(i-1)}\mathbf{S} - \mathbf{S}\mathbf{P}_{(i-1)})$
		
		$\mathbf{K} \leftarrow \mathbf{K} + \max\left(|\mathrm{diag}(\mathbf{K})|\right)\mathbf{I}$
		
		\While{$\mathrm{not\, converged}$}{
			$\mathbf{v}_{i} \leftarrow \frac{\mathbf{K}\mathbf{v}_{i}}{\| \mathbf{K}\mathbf{v}_{i} \|}$
		}
	}
	\caption{Maximum correntropy power iteration for determining the $i$-th principal component.}
\end{algorithm}

\begin{algorithm}[t]
\SetAlgoLined
\SetKwInOut{Input}{Input}
\SetKwInOut{Output}{Output}
\Input{$\mathbf{X}$, $\eta$, $n_{decay}$}
\Output{$\mathbf{V}$}
 $\mathbf{V}_{(0)}, \mathbf{\Lambda} \leftarrow \mathrm{EVD}\left(\frac{1}{n}\mathbf{X}^{T}\mathbf{X}\right)$
 
 $ \mathbf{P}_{(0)} \leftarrow \mathbf{O}$
 
 $ \mathbf{Q}_{(0)} \leftarrow \mathbf{I}$
 
 \For{$i = 1:(p-1)$}{
 	
  $\sigma_{i} \leftarrow \sqrt{\lambda_{i}}$
  
  \For{$j = 1:n_{decay}$}{
  	
   $\mathbf{v}_{i} \leftarrow \mathrm{Alg.\,2}\!\left(\mathbf{X}, \mathbf{P}_{(i-1)}, \mathbf{Q}_{(i-1)}, \mathbf{v}_{i}, \sigma_{i}\right)$
   
   $\sigma_{i} \leftarrow \eta\sigma_{i}$
   
  }
  $\mathbf{P}_{(i)} \leftarrow \mathbf{P}_{(i-1)} + \mathbf{v}_{i}\mathbf{v}_{i}^{T}$
  
  $\mathbf{Q}_{(i)} \leftarrow \mathbf{Q}_{(i-1)} - \frac{ \mathbf{Q}_{(i-1)}\mathbf{v}_{i}\mathbf{v}_{i}^{T}\mathbf{Q}_{(i-1)}}{1 + \mathbf{v}_{i}^{T}\mathbf{Q}_{(i-1)}\mathbf{v}_{i}}$
  
 } 

 $\mathbf{v}_{p} \leftarrow \mathrm{Null}(\mathbf{V}_{1:(p-1)}^{T})$
 \caption{Kernel size shrinking algorithm for determining the entire set of principal components.}
\end{algorithm}

To this end, we propose a sequential procedure, given in Algorithm 3, which gradually shrinks the kernel size, $\sigma$, using a decay factor, $0 < \eta < 1$. In this way, after the $i$-th principal component has been obtained using a fixed kernel size $\sigma$, the procedure is repeated using a smaller kernel size $\eta\sigma < \sigma$. For each $i$-th principal component, we initialise the kernel size, $\sigma_{i}$, as the singular value associated with the $i$-th \textit{a priori} principal component, $\sqrt{\lambda_{i}}$ (obtained from the unweighted scatter matrix $\mathbf{X}^{T}\mathbf{X}$), as it represents a natural upper bound on the value of $\sigma_{i}$.

\begin{remark}
	 The update to the projection matrix $\mathbf{P}_{(i)}$ after a new principal component has been found is of rank-$1$. More generally, we can express the update for the matrix inverse $\mathbf{Q}_{(i)}$ in a simplified form using Woodbury's matrix identity \cite{woodbury}, provided in line $11$ of Algorithm $3$.
\end{remark}

\section{Simulations}
\label{sec:simulations}

The robustness of the proposed maximum correntropy power iteration is now demonstrated through an illustrative and intuitive numerical example. We compared the performance of our maximum correntropy power iteration method with that of standard PCA, applied to a data matrix $\mathbf{X} \in \mathbb{R}^{n \times p}$, with dimensions $n = 400$ and $p = 3$. The data was generated from a $p$-variate Gaussian distribution, ${\bf x}_{k} \sim \mathcal{N}({\bf 0},{\bf S})$, with arbitrarily selected parameters. The true scatter matrix of the covariates was defined as follows
\begin{equation}
    \mathbf{S} := 
    \begin{bmatrix}
        8 & 3 & -1 \\
        3 & 4 & -2 \\
        -1 & -2 & 6
    \end{bmatrix}
\end{equation}
and its eigenvalue decomposition, $\mathbf{S} \equiv \mathbf{V}\mathbf{\Lambda}\mathbf{V}^{T}$, yielded respectively the following matrices of eigenvectors and eigenvalues
\begin{equation}
    \mathbf{V} \simeq 
    \begin{bmatrix}
        -0.78 & 0.51 & 0.37 \\
        -0.49 & -0.11 & -0.87 \\
        0.40 & 0.85 & -0.33
    \end{bmatrix}, \,\, \mathbf{\Lambda} \simeq 
    \begin{bmatrix}
        10.40 & 0 & 0 \\
        0 & 5.66 & 0 \\
        0 & 0 & 1.94
    \end{bmatrix}
\end{equation}
Figure 1(a) shows the scattered data points, together with the directions of the true principal components. In addition, we display the principal components estimated by our maximum correntropy power iteration in Algorithm 3 (with parameters set to $\eta = 0.95$ and $n_{decay} = 65$) and those obtained from standard PCA. As desired, in the absence of outliers or non-Gaussianity in data, both algorithms found the same true principal components.

We also considered a second scenario, where the data matrix, ${\bf X}$, was corrupted with \textit{i.i.d.} outliers, $\boldsymbol{\epsilon}_{k} \sim \mathcal{N}({\bf 0}, \nu {\bf \Lambda})$, with $\nu$ defined as a positive factor to scale the power of the outliers. For this example, we set $\nu=15$. Figure 1(b) shows the same scattered data points, but with $5\%$ of the samples, ${\bf x}_{k}$, replaced with outliers, $\boldsymbol{\epsilon}_{k}$. In this case, observe that while the principal components obtained from standard PCA are misaligned due to the influence of the outliers, the principal components obtained from the proposed maximum correntropy power iteration remain firmly close to the true components.


\begin{figure}[t]
	\vspace{-0.25cm}
	\centering
	\begin{subfigure}[t]{0.5\textwidth}
		\centering
		\includegraphics[width=0.8\textwidth]{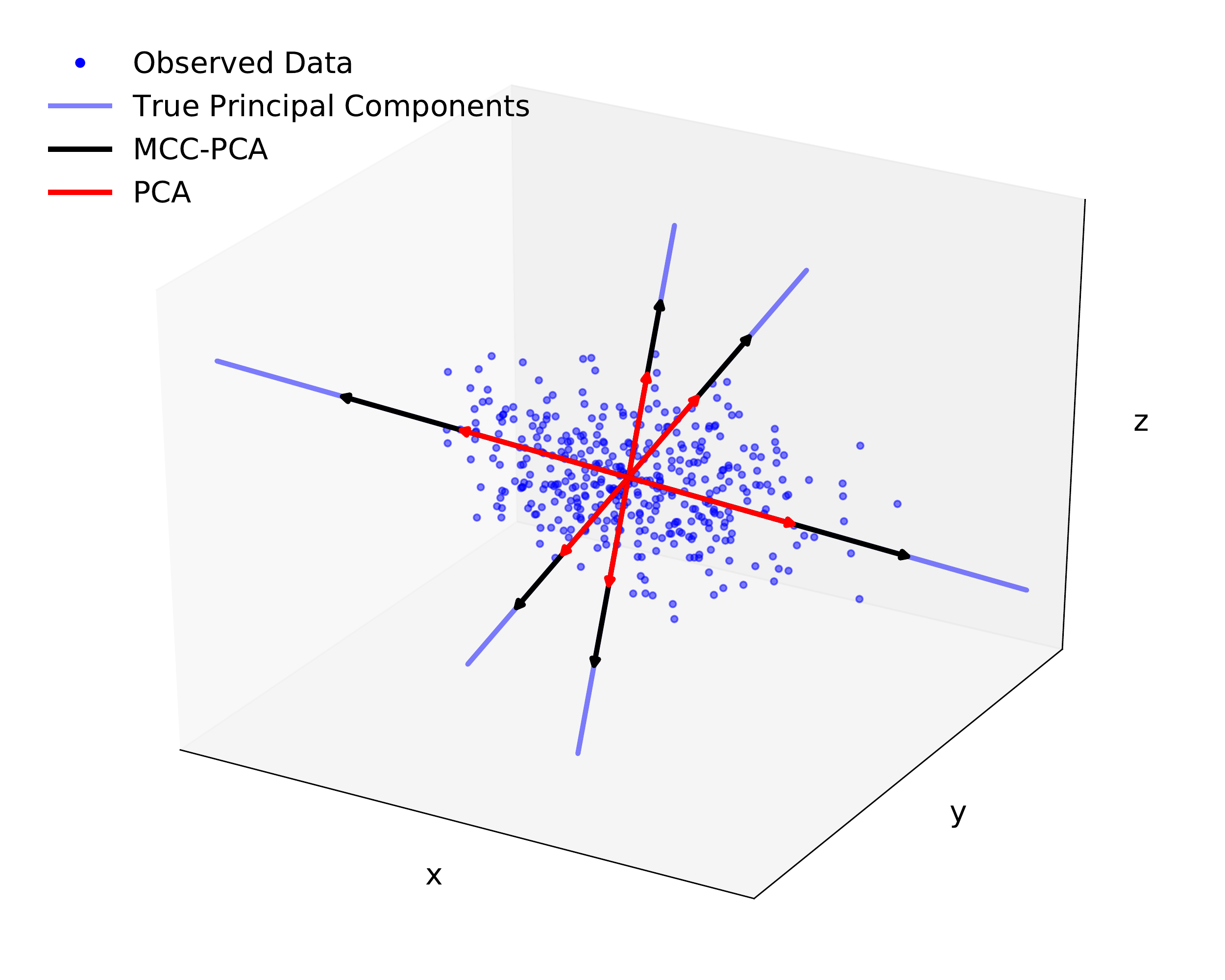} 
		\vspace{-0.25cm}
		\caption{Gaussian distributed samples.}  
	\end{subfigure}
	\hfill
	\begin{subfigure}[t]{0.5\textwidth}   
		\centering 
		\includegraphics[width=0.8\textwidth]{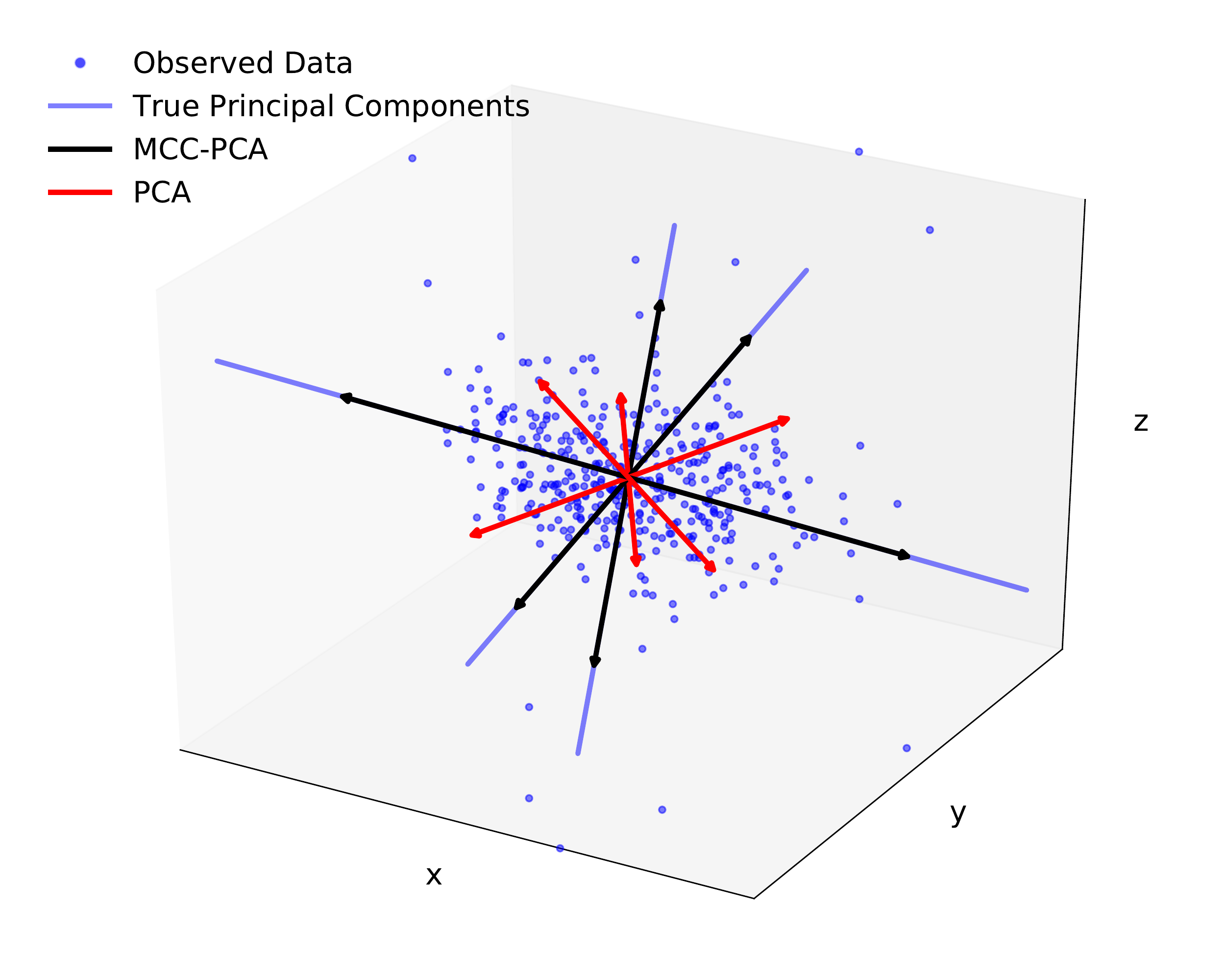} 
		\vspace{-0.25cm}
		\caption{Gaussian distributed samples with $5\%$ of outliers.}
	\end{subfigure}
	\vspace{-0.25cm}
	\caption{Estimates of the principal components based on the proposed maximum correntropy power iteration and standard PCA.}
	\vspace{-0.5cm}
\end{figure}

%

\section{Conclusion}
\label{sec:conclusion}

A robust formulation of principal component analysis (PCA) has been derived from the perspective of the maximum correntropy criterion (MCC), whereby the expected likelihood of the reconstruction error being Gaussian distributed is maximised. In this way, the estimation of principal components becomes insensitive to outliers and non-Gaussian samples. We have shown that the proposed solution, in the form of a generalised power iteration, naturally generalises to any robust M-Estimator based optimisation problem \cite{mestimator}. The proposed maximum correntropy power iteration scheme has been devised so as not to contain costly matrix operations, thereby offering effective means for large-scale applications. Future work aims at examining the convergence of the proposed power iteration scheme and its utility in a wide range of real-world applications.

\pagebreak
\bibliographystyle{IEEEbib}
\bibliography{strings,refs}

\end{document}